\documentclass{article}

\usepackage{fancyvrb}
\usepackage{graphicx}
\usepackage{titling}

\predate{}
\date{}
\postdate{}

\begin{document}

\title{Code Farming: A Process for Creating \protect\\ Generic Computational Building Blocks\protect\footnote{Patent pending.}}
\author{David Landaeta \protect\\ Natural Computation LLC \protect\\
\tt natucomp@gmail.com}
\maketitle

\begin{abstract}

Motivated by a desire to improve on the current state of the art in genetic programming, and aided by recent progress in understanding the computational aspects of evolutionary systems, we describe a process that creates a set of generic computational building blocks for the purpose of seeding initial populations of programs in any genetic programming system.
This provides an advantage over the standard approach of initializing the population purely randomly in that it avoids the need to constantly rediscover such building blocks. It is also better than seeding the initial population with hand-coded building blocks, since it lessens the amount of human intervention required by the system.

\end{abstract}

\section{Introduction} \label{sec:intro}

Genetic programming (GP) is a form of biologically-inspired machine learning that leverages the analogy between computer program code and natural genetic code. We use a mapping between the two domains as follows.
\begin{eqnarray*}
\mbox{genotype} & = & \mbox{program} \\
\mbox{phenotype} & = & \mbox{function encoded by the program} \\
\mbox{allele} & = & \mbox{program fragment} \\
\mbox{gene} & = & \mbox{locus of the program fragment}
\end{eqnarray*}

A GP system may be implemented in a variety of ways, but it must at least specify a choice of programming language, a set of genetic operators for recombining parent programs into child programs, a target problem to solve, and a fitness function that determines how well a given program solves the target problem. Typically, a population of programs is initialized with randomly generated code, then the following steps are repeated until some termination criterion is satisfied, such as obtaining a program that exceeds some threshold of fitness.
\begin{enumerate}
\item Apply the fitness function to produce a score for each program in the population.
\item Select programs according to their score using a probabilistic method, such as {\em roulette wheel selection}~\cite{goldberg}.
\item Apply the genetic operators to the selected programs to produce a new population.
\end{enumerate}

A possible advantage for GP over other forms of machine learning is that it should have the ability to compress information in the same way that the complex functionality of a biological organism is compressed into its DNA. According to Hutter~\cite{hutter}, being able to compress well is closely related to acting intelligently. Moreover, there is some hope that a GP system will be able to explain how it makes decisions in that the program code that it produces can be expressed in a human-readable programming language. Of course, there is no guarantee it will be easy to understand the code, but software engineers routinely maintain code they did not write, which all too often is poorly documented and poorly written, so the necessary skill set for this task is widely available.

By contrast, a deep neural network that is trained through backpropagation typically has enough information storage capacity in its weights and biases to contain its training dataset in an uncompressed form~\cite{zhang}. Even when properly trained, the network is a black box that offers no explanation of how decisions are made. The resulting machine learning systems can be very successful at specific tasks, but they tend to exhibit weaknesses, such as brittleness~\cite{su}, which are not found in human intelligence. There is growing concern that such systems are reaching their useful limits, and cannot achieve the goal of artificial general intelligence without help from other techniques~\cite{marcus}.

\section{The Building Block Hypothesis} \label{sec:bbh}

The building block hypothesis is often used as the justification for why GP is practical~\cite{goldberg}. Intuitively, it doesn't appear to be practical, because searching for a program that solves a given problem by randomly generating programs and trying each one would seem to require time that is exponential in the program size. The building block hypothesis claims that, rather than needing to generate a perfect program, we only need to find relatively short program fragments, or {\em building blocks}, which tend to be present in programs giving good approximations for the solution. By leveraging parallel computation, many building blocks can be found at the same time, and the genetic operators used in GP are designed to combine building blocks into a reasonably good solution.

Assuming the building block hypothesis is true, there is still a problem: even though we only need to spend the time required to find short building blocks, we must find building blocks every time we use GP. This can mean searching for the same building blocks over and over again, which wastes computational resources.

Many GP systems address this problem by seeding the initial population of programs with hand-coded building blocks that are expected to be useful in solving the particular problem under consideration. For example, if the problem is to estimate the distance between objects in a given image, then the population might be seeded with the trigonometric functions and the value of the constant $\pi$. The drawback of this approach is that it relies too much on human intelligence when the goal is to produce {\em artificial}\ intelligence; plus it opens the door to injecting human bias into the system~\cite{friedman}.

We propose instead an automated process for producing building blocks that are generic in the sense that they tend to be useful in solving a wide variety of problems. Such generic building blocks would be used rather than hand-coded building blocks in seeding the initial population of programs. The process for creating them, which we call {\em code farming}, is itself an instance of GP that is given an extremely hard problem to solve. It doesn't matter whether this particular problem can be solved by the GP system; it only matters that, as a byproduct of its attempt to solve the problem, the GP system produces generic building blocks within its population of programs. The process will require large amounts of computing resources in order to produce a significant collection of generic building blocks, but it's worth the effort considering that such building blocks are infinitely reusable, and can thus save resources in the long run.

\begin{figure}
  \includegraphics[height=0.9\textheight]{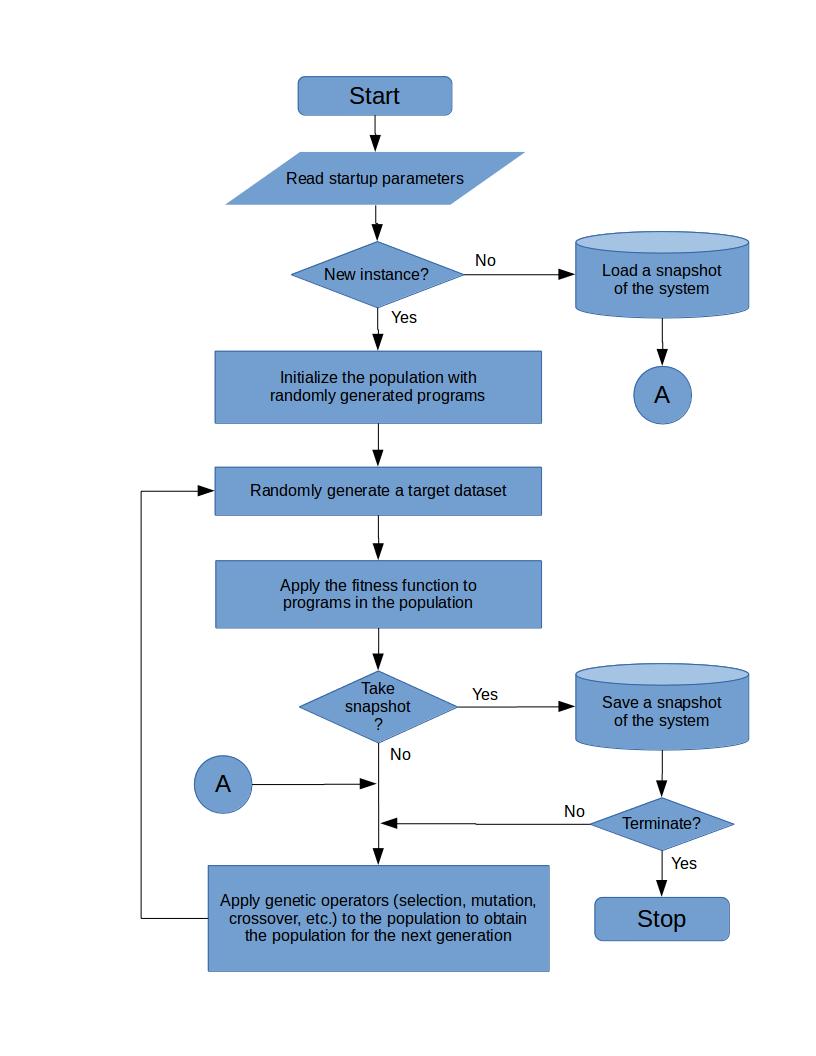}
  \caption{A summary of the process}
  \label{fig:proc}
\end{figure}

\section{The Code Farming Process} \label{sec:proc}

The process is summarized in Figure~\ref{fig:proc}. It is an instance of GP that is distinguished by the fact that it is given an extremely hard problem to solve. The problem in question is to produce a program whose input-output pairs match as closely as possible a given dataset of input-output pairs that is {\em randomly generated}. To make the problem even harder, a new such target dataset is randomly generated with each new generation of the population of programs.

This seems like an absurd task, but recall from Section~\ref{sec:bbh} that we don't actually expect the task to be solved; in fact, it would be a very bad result if the population of programs were to converge on a supposed solution. Instead, we want the population to churn relentlessly, with a variety of possible solutions present at all times.

There is wide latitude on the properties of a GP instance that can successfully implement this process, but the following requirements are essential.
\begin{enumerate}
\item \label{itm:turing} The programming language is {\em Turing complete}, meaning that it can encode any function that a Turing machine can encode~\cite{banzhaf}.
\item \label{itm:sex} The genetic operators implement some form of sexual recombination; for example {\em crossover}~\cite{goldberg}.
\item \label{itm:fitness} The {\em weak selection}\ assumption applies, meaning that the variation in fitness between genotypes is small compared to the recombination rate~\cite{chastain}.
\end{enumerate}
Item~\ref{itm:turing} is necessary in order to ensure the reusability of the building blocks discovered by the process. Items~\ref{itm:sex} and~\ref{itm:fitness} are requirements for the theoretical justification for the process presented in Section~\ref{sec:theory}. Further requirements are described in the following subsections.

\subsection{The End Product} \label{sec:endprod}

What is the end product of this process, and how do we use it?
We said that we expect the process to produce generic building blocks, but these are program fragments. How do we know which fragments within the population of programs are the right ones? It turns out we don't need to identify specific fragments as building blocks; instead, we use entire programs produced by the process. This is because, as long as the correct building blocks are present somewhere in the population, the GP mechanics should be able to find them and combine them together into a good solution.

Whenever a new instance of GP is to be seeded with generic building blocks, its initial population is chosen from the programs that were produced by the code farming process. But which programs in particular do we use? The obvious answer is to choose from the last generation of programs produced. However, we find we get better results by selecting from the collection of programs obtained by taking the most fit program in each generation of the process, with preference given to later generations. That is, with each generation of the process, after the fitness function is applied (see Figure~\ref{fig:proc}), we make a copy of the program with the highest fitness score, where ties are broken arbitrarily, and we add it to the end of a persistent list of programs, called the {\em seed list}, which is initially empty. If the new GP instance requires an initial population of size $n$, then it uses the last $n$ entries on the seed list. Thus, the seed list is the end product of the process.

A slight variation of this approach applies a filter to the seed list to remove duplicate phenotypes before using the list to initialize a population. This is useful when we want to keep the size of the population small. The technique of using {\em function signatures}, as described in the next section, can be used to detect duplicate phenotypes in a probabilistic sense.

\subsection{Elitism} \label{sec:elitism}

A well-known problem with GP is that a randomly generated program in any sufficiently rich programming language---especially one that is Turing complete---will tend to encode a function that is trivial, such as a function that always ignores its input and produces constant output. This means that the GP system might never get {\em aloft}; in other words, it might never produce interesting behavior. Furthermore, even if non-trivial behavior is produced, it tends to be fragile, in that it is easily destroyed by genetic operators, so the system might not stay aloft. In theory, the probability of getting and staying aloft can be increased by maintaining a sufficiently large population of programs, but this might require vast computational resources. So as a practical matter, we need to provide an efficient mechanism for getting and staying aloft.

We find one such mechanism to be particularly effective in the case of the code farming process, namely {\em elitism}~\cite{goldberg}. Under this approach, certain highly fit individuals, the so-called {\em elites}, are given added assurance of surviving from one generation to the next. We use a variation of elitism with the following properties.
\begin{enumerate}
\item The list of elites is persistent over all generations and is initially empty.
\item With each generation, after the fitness function is applied, a single individual is added to the elites if and only if it satisfies the following.
\begin{enumerate}
\item It is the fittest individual in the current generation, where ties are broken arbitrarily.
\item The function it encodes is distinct from that encoded by any current elite. This is determined efficiently in a probabilistic sense by randomly generating a persistent list of test inputs during system initialization, and declaring a candidate function to be distinct if and only if it has a distinct sequence of outputs for the test inputs. We refer to this sequence of outputs as the function's {\em signature}.
\end{enumerate}
\item In order to provide some assurance that non-trivial functions are added to the elite list at all, the fitness function is modified to flag obvious cases of programs encoding trivial functions, which are then assigned the lowest possible score. For example, in many cases, it is possible for an automated syntactic analysis of the program to determine that the input will never be read, so the program can be flagged as trivial without ever running it. Note, however, the fact that our chosen programming language is Turing complete implies there is no algorithm for determining in all cases whether a given program ever reads its input~\cite{rogers}.
\item The advantage given to elites is that with a small probability, called the {\em elite probability}, the selection of an individual to be a parent for the next generation comes from the elite list rather than the current generation, and in that case every elite has equal probability of being selected.
\end{enumerate}

\subsection{Measuring Progress} \label{sec:progress}

How do we know whether the process is making progress, and how do we know when enough progress has been made to terminate the process? We find the best answer to both questions is based on the size of the elite list: if it's growing at a steady pace, then steady progress is being made, and there should be at least a few thousand elites found before considering termination of the process, but more is better.

A more reliable---but more resource-intensive---approach is to periodically use copies of the seed list to initialize populations within a test suite of GP instances representing a variety of problems, and determining if the test results are improving over time. If the test results provide acceptable solutions, then the process can be terminated.

\subsection{Snapshots} \label{sec:snapshots}

As shown in Figure~\ref{fig:proc}, the process has steps that load or save a {\em snapshot}\ of the system, which consists of all the data needed to allow the process to pause and resume at a later time. A snapshot must contain at least the following.
\begin{enumerate}
\item The current population of programs and their fitness scores.
\item The seed list.
\item The list of elites and their signatures.
\item The list of test inputs for determining signatures.
\end{enumerate}

\subsection{Target Datasets} \label{sec:datasets}

A target dataset is randomly generated with each new generation of the process. It consists of input-output pairs, but what precisely are their formats?
In order to answer this question, we must make a design trade-off between the degree to which the end product is reusable in various contexts and the speed and efficiency with which it can be produced.

If we want the end product to be fully reusable for a wide range of problems, then the input-output pairs should be pairs of binary strings having no length restriction, apart from being finite, since such strings can encode any information whatever. In this case, there is a probability density function, which is a variation of the well-known uniform density function, that clarifies what it means to randomly generate a binary string of arbitrary length~\cite{li}. The function is given by
\[
L(x) = 2^{-2 l(x) - 1}
\]
where $x$ is any binary string of length $l(x)$.

However, we find that the lack of restrictions on input-output pairs requires increased computational resources in order to produce a usable end product. It is much more efficient to focus on a particular problem domain, and restrict input-output pairs to a format that is appropriate for that domain while still being randomly generated. For example, in Zhang {\em et al.}~\cite{zhang}, a random dataset that is appropriate for the problem domain of image classification is generated using pure Gaussian noise as input and uniformly random labels as output.

\subsection{The Fitness Function} \label{sec:fitness}

The fitness function must measure how well the function encoded by a given program matches the current target dataset. It is also required to satisfy the weak selection assumption, which, according to Chastain {\em et al.}~\cite{chastain}, means that
the fitnesses of all genotypes (programs) are
close to one another, say within the interval $[1 - \varepsilon, 1 + \varepsilon]$, and so
the fitness of genotype $g$ can be written as $F_g = 1 + \varepsilon \Delta_g$, where
$\varepsilon$ is
the {\em selection strength}, assumed to be small, and
$\Delta_g \in [-1, 1]$
is called the {\em differential fitness}\ of the genotype.
These requirements are easily satisfied by defining $\Delta_g$ to be what one would normally consider to be the fitness score; for example, $\Delta_g = -1$ means that $g$ provides the worst match for the target dataset relative to other members of the population, and $\Delta_g = 1$ means that $g$ provides the best match relative to others.

Note that there is some flexibility in the definition of ``match.'' In many cases, it is useful to loosen the definition to mean {\em highly correlated}\ rather than {\em exact match}, since this increases the chances of finding an acceptable solution. For instance, we might consider a program that produces the exact opposite (in a bitwise sense) of the target output to be as fit as an exact match, given that it is easily transformed into a perfect solution.

\section{Experimental and Theoretical Justification} \label{sec:theory}

It might seem pointless to run a GP system whose objective changes randomly with every generation. It looks like we're shooting at a moving target, with no hope of success. Our intuition tells us that the population will never amount to anything more than the random jumble of code with which it starts. However, recent theoretical work by Livnat
{\em et al.}~\cite{livnat} provides justification for the process.

That work reflects a growing trend in the analysis of the role of sex in evolution. The new thinking is that, when sexual recombination is present, evolution produces high-quality {\em alleles}\ rather than high-quality individuals, as is traditionally thought. This explains why we need not be concerned about the fact that the code farming process cannot possibly produce an individual program that is consistently successful at matching randomly generated target datasets. However, it's still not clear how one allele (code fragment) can be better than another at this task. We claim that the addition of the weak selection assumption makes this possible, and in the following subsections we provide experimental and theoretical justification for this claim.

\begin{figure}
\centering
\caption{Typical output from program {\tt demo.cc}}
\label{fig:output}
\begin{BVerbatim}[fontsize=\small]


Generation Allele:0 Allele:1
---------- -------- --------
         0      50%      49%
        20      41%      58%
        40      31%      68%
        60      26%      73%
        80      39%      60%
       100      40%      59%
       120      36%      63%
       140      51%      48%
       160      55%      44%
       180      62%      37%
       200      63%      36%
       220      47%      52%
       240      50%      49%
       260      18%      81%
       280      14%      85%
       300       4%      95%
       320      11%      88%
       340       8%      91%
       360       5%      94%
       380      20%      79%
       400      30%      69%
       420      34%      65%
       440      59%      40%
       460      63%      36%
       480      67%      32%
       500      70%      29%
       520      76%      23%
       540      53%      46%
       560      72%      27%
       580      70%      29%
       600      31%      68%
       620      50%      49%
       640      23%      76%
       660      28%      71%
       680      23%      76%
       700      20%      79%
       720      17%      82%
       740      13%      86%
       760      12%      87%
       780      12%      87%
       800      25%      75%
---------- -------- --------
   Average      37%      62%
\end{BVerbatim}
\end{figure}

\subsection{A Simple Experiment} \label{sec:experiment}

We provide a simple, reproducible experiment to demonstrate that there exists a GP system for which a fitness function based on matching a randomly-generated target dataset consistently favors one type of allele over another. A complete, self-contained C++ code listing for running the experiment is given in Appendix~\ref{app:demo}.
This GP system is a highly simplified analog of the code farming process. It does not contain features that are inessential for the demonstration; for example, the programming language is not Turing complete, and the system does not use elitism.

A genotype in this system is a sequence of $k + 1$ bits, where each bit is a gene with alleles 0 and~1. The first gene is a master control that determines how the remaining $k$ genes are used in the function $\phi$ that is encoded, which takes as input a sequence of $2^k$ bits and produces a single bit as output. Allele 0 of the control gene ignores the remaining genes and defines $\phi$ to be the function that returns 0 for every input, whereas allele 1 interprets the $k$ bits of the remaining genes as an unsigned integer representing a position in the input sequence, and $\phi$ returns the value of the input bit in that position as its output.

Figure~\ref{fig:output} shows typical output from a single run of the experiment. Multiple runs show a clear advantage for allele~1 of the control gene over allele~0. The next subsection gives a theoretical explanation for this result.

\subsection{Nagylaki's Theorem} \label{sec:nagylaki}

Nagylaki's theorem~\cite{nagylaki} describes how the frequency of an allele changes over time, assuming that sexual recombination and weak selection are both present:
\[
x_{i}^{t+1}(j) = \frac{F_{i}^{t}(j)}{X^t} x_{i}^{t}(j),
\]
where
\begin{itemize}
\item $x_{i}^{t}(j)$
is the frequency in the population of allele $j$ of locus $i$ at generation $t$,
\item $F_{i}^{t}(j)$ is the mean fitness at generation $t$ over all genotypes that contain allele $j$ at locus $i$, and
\item $X^t$ is a normalizing constant designed to keep the frequencies summing to 1 at generation $t$.
\end{itemize}

This theorem has an interesting property: in cases where two alleles of the same locus have identical time-averaged mean fitness, it favors a consistent performer over an erratic performer. In other words, it favors the allele with the smaller variance in mean fitness over time. This is a straightforward consequence of the inequality of arithmetic and geometric means~\cite{steele}.
\pagebreak

But note that when the fitness function is based on matching a randomly-generated target dataset, and the target dataset is randomly recreated with each generation of the population, then {\em all alleles have identical time-averaged mean fitness}.
So if this fitness function is used in a GP system, together with sexual recombination and weak selection, then the only possible way for one allele to be more successful than another is by having a smaller variance in mean fitness over time.

The GP system of Section~\ref{sec:experiment} shows exactly how this can occur. Allele 0 of the control gene always corresponds to only a single phenotype, whereas allele 1 corresponds to a set $\Phi = \{ \phi_1, \ldots, \phi_k \}$ of $k$ distinct phenotypes.
Moreover, since the functions in $\Phi$ are mutually uncorrelated, and since the inputs and outputs in the dataset are randomly chosen, the set of fitness values of $\Phi$ can be viewed as a set of mutually independent and identically distributed discrete random variables. By the law of large numbers~\cite{hoel}, the mean fitness of $\Phi$ at time $t$ should get closer to the time-averaged mean fitness of $\Phi$ as $k$ gets larger. Thus, for sufficiently large $k$, the variance in mean fitness over time of allele 1 will be lower than that of allele 0 with high probability.

\section{Conclusion} \label{sec:conclusion}

In the previous section, we demonstrated conditions under which one allele can be more successful than another at the task of matching a randomly-generated target dataset. We now claim that these conditions for success are the same criteria that define a generic computational building block.

In Section~\ref{sec:bbh}, we defined a generic computational building block to be a code fragment that tends to be useful in solving a wide variety of problems. In terms of genetics, this corresponds to an allele that tends to be present in a family of genotypes encoding a wide variety of distinct phenotypes. If we interpret ``variety'' to mean the same thing as ``mutually independent,'' which is a reasonable definition, then we arrive at the same condition for an allele to be successful at matching random datasets, as required by the claim.

This concept of a generic computational building block is strongly related to the concept of {\em modularity}, which occurs throughout a wide range of fields~\cite{baldwin}. More research is needed in order to precisely describe this relationship.

\appendix
\section{\tt demo.cc} \label{app:demo}

\begin{Verbatim}[fontsize=\small]
/*
 * A C++11 program to demonstrate that there exists a GP system for which
 * a fitness function based on matching a randomly-generated target dataset
 * consistently favors one type of allele over another.
 */

#include <algorithm>
#include <cassert>
#include <cstdlib>
#include <functional>
#include <iomanip>
#include <iostream>
#include <random>
#include <vector>

typedef std::vector<bool> Input;
typedef bool Output;
typedef struct { Input input; Output output; } Datum;
typedef std::vector<Datum> Dataset;
typedef std::function<Output(const Input&)> Phenotype;
typedef std::vector<bool> Genotype;
typedef std::vector<Genotype> Population;
typedef std::vector<double> Scores;

enum class CrossoverMethod {
    kSingle = 0,
    kUniform = 1
};

struct Configuration {
    std::mt19937 *random_engine_ptr;
    CrossoverMethod crossover_method;
    double mutation_rate;
    double crossover_rate;
    double selection_strength;
    int num_genes;
    int num_inputs;
    int num_genotypes;
    int num_examples;
    int num_generations;
    int report_interval;
};

void Mutate(
        const Configuration& c,
        Genotype *genotype_ptr) {
    std::bernoulli_distribution mutation(c.mutation_rate);
    for (auto&& bit : *genotype_ptr) {
        bit = bit != mutation(*c.random_engine_ptr);
    }
}

void SingleCrossover(
        const Configuration& c,
        Genotype *genotype_1_ptr,
        Genotype *genotype_2_ptr) {
    std::bernoulli_distribution crossover(c.crossover_rate);
    if (!crossover(*c.random_engine_ptr)) {
        return;
    }
    auto&& genotype_1 = *genotype_1_ptr;
    auto&& genotype_2 = *genotype_2_ptr;
    std::uniform_int_distribution<int> crossover_index(
            1, std::max(1, c.num_genes - 1));
    int index = crossover_index(*c.random_engine_ptr);
    for (int i = index; i < c.num_genes; ++i) {
        Genotype::swap(genotype_1[i], genotype_2[i]);
    }
}

void UniformCrossover(
        const Configuration& c,
        Genotype *genotype_1_ptr,
        Genotype *genotype_2_ptr) {
    std::bernoulli_distribution crossover(c.crossover_rate);
    auto&& genotype_1 = *genotype_1_ptr;
    auto&& genotype_2 = *genotype_2_ptr;
    for (int i = 0; i < c.num_genes; ++i) {
        if (crossover(*c.random_engine_ptr)) {
            Genotype::swap(genotype_1[i], genotype_2[i]);
        }
    }
}

void Crossover(
        const Configuration& c,
        Genotype *genotype_1_ptr,
        Genotype *genotype_2_ptr) {
    switch(c.crossover_method) {
    case CrossoverMethod::kSingle:
        SingleCrossover(c, genotype_1_ptr, genotype_2_ptr);
        break;
    case CrossoverMethod::kUniform:
        UniformCrossover(c, genotype_1_ptr, genotype_2_ptr);
        break;
    default:
        break;
    }
}

void GeneticOperators(
        const Configuration& c,
        const Scores& scores,
        const Population& old_population,
        Population *new_population_ptr) {
    Population& new_population = *new_population_ptr;
    std::discrete_distribution<int> selection(scores.begin(), scores.end());
    for (int i = 0; i < c.num_genotypes - 1; i += 2) {
        Genotype *genotype_1_ptr = &new_population[i];
        Genotype *genotype_2_ptr = &new_population[i + 1];
        *genotype_1_ptr = old_population[selection(*c.random_engine_ptr)];
        *genotype_2_ptr = old_population[selection(*c.random_engine_ptr)];
        Mutate(c, genotype_1_ptr);
        Mutate(c, genotype_2_ptr);
        Crossover(c, genotype_1_ptr, genotype_2_ptr);
    }
}

void InitializePopulation(
        const Configuration& c,
        Population *population_ptr) {
    std::bernoulli_distribution coin_flip(0.5);
    for (auto&& genotype : *population_ptr) {
        for (auto&& bit : genotype) {
            bit = coin_flip(*c.random_engine_ptr);
        }
    }
}

void GenerateDataset(
        const Configuration& c,
        Dataset *dataset_ptr) {
    std::bernoulli_distribution coin_flip(0.5);
    for (auto&& datum : *dataset_ptr) {
        for (auto&& bit : datum.input) {
            bit = coin_flip(*c.random_engine_ptr);
        }
        datum.output = coin_flip(*c.random_engine_ptr);
    }
}

Phenotype Decode(
        const Configuration& c,
        const Genotype& genotype) {
    if (genotype[0]) {
        int index = 0;
        for (int i = 1; i < c.num_genes; ++i) {
            index <<= 1;
            index |= genotype[i];
        }
        return [index](const Input& input) { return input[index]; };
    } else {
        return [](const Input&) { return false; };
    }
}

void Fitness(
        const Configuration& c,
        const Population& population,
        const Dataset& dataset,
        Scores *scores_ptr) {
    Scores& scores = *scores_ptr;
    // compute raw scores
    for (int i = 0; i < c.num_genotypes; ++i) {
        auto&& phenotype = Decode(c, population[i]);
        double& score = scores[i];
        score = 0.0;
        for (auto&& datum : dataset) {
            score += (datum.output == phenotype(datum.input)) ? 1 : -1;
        }
    }
    // scale scores so that min --> -1.0 and max --> 1.0
    double a = 0.0;
    double b = 0.0;
    double min = *std::min_element(scores.begin(), scores.end());
    double max = *std::max_element(scores.begin(), scores.end());
    if (min != max) {
        a = 2.0 / (max - min);
        b = 1.0 - a * max;
    }
    for (double& score : scores) {
        score = a * score + b;
    }
    // enforce weak selection
    for (double& score : scores) {
        score = 1.0 + score * c.selection_strength;
    }
}

void ReportHeader() {
    std::cout << "Generation Allele:0 Allele:1" << std::endl
              << "---------- -------- --------" << std::endl;
}

void Report(
        const Configuration& c,
        const Population& population,
        int generation,
        int *allele_0_sum_ptr,
        int *allele_1_sum_ptr,
        int *count_ptr) {
    int allele_0 = 0;
    int allele_1 = 0;
    for (auto&& genotype : population) {
        if (genotype[0]) {
            ++allele_1;
        } else {
            ++allele_0;
        }
    }
    *allele_0_sum_ptr += allele_0;
    *allele_1_sum_ptr += allele_1;
    ++*count_ptr;
    std::cout << std::setw(10) << generation << " " << std::setw(7)
              << (100 * allele_0 / c.num_genotypes) << "% "
              << std::setw(7)
              << (100 * allele_1 / c.num_genotypes) << "%"
              << std::endl;
}

void ReportFooter(
        const Configuration& c,
        int allele_0_sum,
        int allele_1_sum,
        int count) {
    int divisor = count * c.num_genotypes;
    std::cout << "---------- -------- --------" << std::endl
              << "   Average " << std::setw(7)
              << (100 * allele_0_sum / divisor) << "% "
              << std::setw(7)
              << (100 * allele_1_sum / divisor) << "%"
              << std::endl;
}

int main() {

    std::random_device random_seeder;
    std::mt19937 random_engine(random_seeder());

    Configuration c{
        /* random_engine_ptr  */ &random_engine,
        /* crossover_method   */ CrossoverMethod::kSingle,
        /* mutation_rate      */ 1.0 / (1 << 11),
        /* crossover_rate     */ 1.0 / (1 << 1),
        /* selection_strength */ 1.0 / (1 << 3),
        /* num_genes          */ 21,
        /* num_inputs         */ 1 << 20,
        /* num_genotypes      */ 1 << 10,
        /* num_examples       */ 1,
        /* num_generations    */ 800,
        /* report_interval    */ 20
    };

    assert(c.mutation_rate >= 0.0 && c.mutation_rate <= 1.0);
    assert(c.crossover_rate >= 0.0 && c.crossover_rate <= 0.5);
    assert(c.selection_strength > 0.0);
    assert(c.num_genes > 1);
    assert(c.num_inputs == (1 << (c.num_genes - 1)));
    assert(c.num_genotypes > 1);
    assert(c.num_examples > 0);
    assert(c.num_generations >= 0);
    assert(c.report_interval > 0);

    Scores scores(c.num_genotypes, 0.0);
    Dataset dataset(
            c.num_examples, Datum{ Input(c.num_inputs), Output(false) });
    std::vector<Population> populations(
            2, Population(
                    c.num_genotypes, Genotype(c.num_genes)));
    int allele_0_sum = 0;
    int allele_1_sum = 0;
    int count = 0;
    int generation = 0;
    int population_toggle = 0;

    InitializePopulation(c, &populations[population_toggle]);
    ReportHeader();

    while (true) {
        population_toggle = 1 - population_toggle;
        auto&& new_population = populations[population_toggle];
        auto&& old_population = populations[1 - population_toggle];
        if (generation % c.report_interval == 0) {
            Report(c, old_population, generation,
                    &allele_0_sum, &allele_1_sum, &count);
        }
        if (generation >= c.num_generations) {
            break;
        }
        ++generation;
        GenerateDataset(c, &dataset);
        Fitness(c, old_population, dataset, &scores);
        GeneticOperators(c, scores, old_population, &new_population);
    }

    ReportFooter(c, allele_0_sum, allele_1_sum, count);

    return EXIT_SUCCESS;
}
\end{Verbatim}

\end{document}